%% file: root.tex
\documentclass[letterpaper, 10 pt, conference]{ieeeconf}  % Comment this line out if you need a4paper

\IEEEoverridecommandlockouts                              % This command is only needed if 
\usepackage{graphics} % for pdf, bitmapped graphics files
\usepackage{amsmath} % assumes amsmath package installed
\usepackage{amssymb}  % assumes amsmath package installed
\usepackage{cite}
\usepackage[ruled, norelsize, linesnumbered]{algorithm2e}
\usepackage{pbox}
\makeatletter
\newcommand{\removelatexerror}{\let\@latex@error\@gobble}
\makeatother
\SetNlSty{text}{}{}
\SetKw{Break}{break}

\usepackage{url}
\usepackage{hyperref}
\hypersetup{colorlinks = true, linkcolor = black, filecolor = black, citecolor = black, urlcolor = blue}

\usepackage[caption=false,font=footnotesize]{subfig}

\usepackage{tikz}
\usepackage{pgfplots}
\pgfplotsset{compat = 1.15}
\usepackage{siunitx}
\usepackage{xspace}
\usepackage{ACIN_colors}
\usepackage{import}
\usepackage{lipsum}

\newtheorem{remark}{Remark}

\graphicspath{graphics}

\DeclareMathOperator{\sign}{sign}

\providecommand{\FullStop}{\text{~\@.\xspace}}
\providecommand{\Comma}{\text{~,\xspace}}

\providecommand{\ROS}{\textsc{ROS}\xspace}

\providecommand{\Optitrack}{\textsc{OptiTrack}\xspace}

\providecommand{\kuka}{\textsc{KUKA} LBR iiwa 14 R820\xspace}

\newcommand{\vect}[1]{\boldsymbol{\mathbf{#1}}}
\newcommand{\dist}[1]{\left\lVert #1 \right\rVert}

\newcommand\scaleddot{\scalebox{.89}{.}}

\makeatletter
\renewcommand{\dddot}[1]{%
  {\mathop{\kern\z@#1}\limits^{\makebox[0pt][c]{\vbox to-1.4\ex@{\kern-\tw@\ex@
   \hbox{\normalfont ...}\vss}}}}}
\newcommand{\dddotsuper}[1]{%
  {\mathop{\kern\z@#1}\limits^{\makebox[0pt][c]{\vbox to-1.6\ex@{\kern-\tw@\ex@
   \hbox{\scriptsize ...}\vss}}}}}
\newcommand{\dddotalt}[1]{%
  {\mathop{\kern\z@#1}\limits^{\makebox[0pt][c]{\vbox to-2.2\ex@{\kern-\tw@\ex@
   \hbox{\normalfont\scaleddot\kern-0.5pt\scaleddot\kern-0.5pt\scaleddot}\vss}}}}}
\makeatother

\providecommand{\qdot}{\dot{\vect{q}}}

\makeatletter
\newcommand{\nosemic}{\renewcommand{\@endalgocfline}{\relax}}% Drop semi-colon ;
\newcommand{\dosemic}{\renewcommand{\@endalgocfline}{\algocf@endline}}% Reinstate semi-colon ;
\newcommand{\pushline}{\Indp}% Indent
\newcommand{\popline}{\Indm\dosemic}% Undent
\makeatother

\DeclareRobustCommand\circled[1]{\tikz[baseline=(char.base)]{
    \node[shape=circle,draw,inner sep=1pt] (char) {\footnotesize #1};}}

\usepackage[utf8]{inputenc}              
\UseRawInputEncoding
\title{\LARGE \bf
Model Predictive Trajectory Optimization With \\Dynamically Changing Waypoints for Serial Manipulators 
}

\author{Florian~Beck$^{1}$, 
        Minh~Nhat~Vu$^{1,2}$, 
        Christian~Hartl-Nesic$^{1}$,
        and~Andreas~Kugi$^{1, 2}$,~\IEEEmembership{Senior Member,~IEEE}% <-this % stops a space
\thanks{$^{1}$F. Beck, M. N. Vu, C. Hartl-Nesic, and A. Kugi are with the Automation and Control Institute, Technische Universit\"at Wien (TUW), 1040 Vienna, Austria (e-mail: \tt\small beck@acin.tuwien.ac.at, vu@acin.tuwien.ac.at, hartl@acin.tuwien.ac.at,\newline kugi@acin.tuwien.ac.at)}%
\thanks{$^{2}$A. Kugi and M. N. Vu are with the AIT Austrian Institute of Technology GmbH, 1210 Vienna, Austria (e-mail: \tt\small Andreas.Kugi@ait.ac.at, Minh.Vu@ait.ac.at)}%
}

\begin{document}

\maketitle
\thispagestyle{empty}
\pagestyle{empty}

%%%%%%%%%%%%%%%%%%%%%%%%%%%%%%%%%%%%%%%%%%%%%%%%%%%%%%%%%%%%%%%%%%%%%%%%%%%%%%%%
\begin{abstract}
%Sophisticated manipulation tasks for serial manipulators require task and motion planning with online replanning capabilities for dynamic environments.
%Furthermore, dynamic environments and partial observations demand online replanning capabilities.
%However, including discrete actions resulting from task planning has been challenging for online model predictive trajectory optimization with short planning horizons.
%Including discrete actions resulting from task planning has been challenging for online model predictive trajectory optimization with short planning horizons. 
%This paper presents the waypoint model predictive trajectory optimization (wMPC) for online replanning tasks with a manipulator, which addresses this issue.
%wMPC considers discrete actions as waypoints in the robot's task space and uses objective functions to plan toward these waypoints. 
%The main contribution is using a constraint to split the planning horizon at the waypoint when it becomes reachable within the current planning horizon. This enables the continuation of planning toward the next waypoint. 
%This provides flexibility in adapting to changing conditions online.
%The results of the presented approach have shown not only competitive path lengths and trajectory durations compared to offline RRT-type planners in a multi-waypoint scenario but also the ability to execute replanning tasks online on a \kuka robot in a dynamic pick-and-place scenario.
Systematically including dynamically changing waypoints as desired discrete actions, for instance, resulting from superordinate task planning, has been challenging for online model predictive trajectory optimization with short planning horizons. This paper presents a novel waypoint model predictive control (wMPC) concept for online replanning tasks. The main idea is to split the planning horizon at the waypoint when it becomes reachable within the current planning horizon and reduce the horizon length towards the waypoints and goal points. This approach keeps the computational load low and provides flexibility in adapting to changing conditions in real-time. The presented approach achieves competitive path lengths and trajectory durations compared to (global) offline RRT-type planners in a multi-waypoint scenario. Moreover, the ability of wMPC to dynamically replan tasks online is experimentally demonstrated on a \kuka robot in a dynamic pick-and-place scenario.
\end{abstract}

%%%%%%%%%%%%%%%%%%%%%%%%%%%%%%%%%%%%%%%%%%%%%%%%%%%%%%%%%%%%%%%%%%%%%%%%%%%%%%%%
\section{INTRODUCTION}
\label{sec:introduction}
    \input{sections/introduction.tex}

    \section{RELATED WORK}
    \label{sec:related_work}
    \input{sections/related_work.tex}

    \section{MATHEMATICAL MODEL}
    \label{sec:math_model}
    \input{sections/model.tex}
    
    \section{WAYPOINT MPC}
    \label{sec:rhtp}
    \input{sections/rhtp.tex}
    
    \section{SIMULATION AND EXPERIMENTAL RESULTS}
    \label{sec:results}
    \input{sections/results.tex}

    \section{CONCLUSIONS}
    \label{sec:conclusion}
    \input{sections/conclusion.tex}

%\addtolength{\textheight}{-0.6cm}   % This command serves to balance the column lengths
                                  % on the last page of the document manually. It shortens
                                  % the textheight of the last page by a suitable amount.
                                  % This command does not take effect until the next page
                                  % so it should come on the page before the last. Make
                                  % sure that you do not shorten the textheight too much.

%%%%%%%%%%%%%%%%%%%%%%%%%%%%%%%%%%%%%%%%%%%%%%%%%%%%%%%%%%%%%%%%%%%%%%%%%%%%%%%%

%%%%%%%%%%%%%%%%%%%%%%%%%%%%%%%%%%%%%%%%%%%%%%%%%%%%%%%%%%%%%%%%%%%%%%%%%%%%%%%%

%%%%%%%%%%%%%%%%%%%%%%%%%%%%%%%%%%%%%%%%%%%%%%%%%%%%%%%%%%%%%%%%%%%%%%%%%%%%%%%%
%\section*{APPENDIX}
%
%Appendixes should appear before the acknowledgment.
%
%\section*{ACKNOWLEDGMENT}

%%%%%%%%%%%%%%%%%%%%%%%%%%%%%%%%%%%%%%%%%%%%%%%%%%%%%%%%%%%%%%%%%%%%%%%%%%%%%%%%

\bibliographystyle{IEEEtran}
\bibliography{references}

\end{document}

%% file: sections/introduction.tex
%\begin{itemize}
%  \item Complex robotic manipulation sequences, especially long-horizon tasks~\cite{Mees2022}
%  \item Task and motion planning
%  \item waypoints motivation
%  \item Problems with discrete waypoints and gradient-based methods
%  \item proposed solution and novelty statement
%\end{itemize}
%
%\lipsum[1-4]

Tasks for robotic manipulators in unstructured human environments demand sophisticated planning techniques.
The dynamic nature and incomplete measurements in such environments require online replanning capabilities to ensure proper execution.
Planning for such tasks can be roughly classified into a discrete sequence of actions to be executed by the robot, referred to as task planning, and planning the robot's motion to complete such actions, i.e., motion planning~\cite{Garrett2021}.
This work considers discrete actions that can be abstracted by waypoints in the robot's task space, e.g., moving to an object to grasp it from a specific pre-grasp point.
%This work considers discrete actions as waypoints in the robot's task space.
%Combined task and motion planning~(TAMP) has made significant progress over the years, acknowledging the interdependence of the two planning problems~\cite{Garrett2021}.
%One category of TAMP is using optimization-based approaches combining trajectory optimization and modeling the discrete action sequences as constraints~\cite{Toussaint2015, Hadfield-Menell2016}.
In trajectory optimization, action sequences or waypoints can be modeled as constraints~\cite{Toussaint2015}.
%Discrete action sequences can be modeled as constraints in trajectory optimization~\cite{Toussaint2015}.
This requires trajectory optimization over a long planning horizon that covers the action sequence's length.
Such an optimization procedure is computationally expensive and, hence, unsuitable for environments where conditions change dynamically, requiring online replanning.
Consider, for example, picking an object and placing it in a cabinet. 
Depending on the available sensors, the robot may not detect whether the cabinet is already open or whether there is any space left in its initial state when starting to plan.
In this case, new observations that become available when approaching the cabinet with an object can require putting the object down and opening the cabinet door before placing the object.

\begin{figure}[t]
  \centering
  \includegraphics{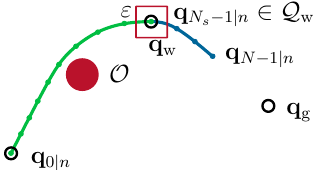}
  \caption{\label{fig:wmpc_illustration} The proposed wMPC planner first plans towards the waypoint $\vect{q}_{\mathrm{w}}$, avoiding the obstacle $\mathcal{O}$. The planner splits the horizon at $k = N_s$ as soon as the waypoint $\vect{q}_{\mathrm{w}}$ is reachable within a tolerance $\varepsilon$. Then, the waypoint is constrained by the planner with $\vect{q}_{N_s - 1 | n} \in \mathcal{Q}_{\mathrm{w}}$ to be within the tolerance band around the waypoint $\vect{q}_{\mathrm{w}}$, and the remaining samples are used to optimize towards the goal point $\vect{q}_{\mathrm{g}}$.}
\end{figure}

A typical approach for online trajectory optimization is model predictive control~(MPC)~\cite{GhazaeiArdakani2019, Kraemer2020, Schoels2020a} over a short, receding horizon.
As discussed before, complex manipulation tasks are often divided into discrete actions obtained from a task planner.
However, it is not apparent how to systematically include discrete-time constraints, such as waypoints, in a receding horizon concept, as these waypoints might only be reachable in future iterations.
Furthermore, the timing of such waypoints is challenging to obtain.
Current attempts to solve this problem rely on a tracked reference path or trajectory to maintain this global view of discrete constraints~\cite{Toussaint2022, Romero2022}.
The disadvantage of such approaches is the need to compute such a reference.
Due to the online requirement, only simplified reference paths or trajectories can be computed, i.e., collision checking is typically neglected.
%Moreover, if significant deviations from this reference are necessary to compute a collision-free trajectory online, the reference can introduce additional local minima.
The approach presented in this work alleviates the requirement of global references for including waypoints in model predictive trajectory optimization with short horizons.
Only the waypoints themselves are needed as inputs to the planner.
The main point is that the objective function is used to plan towards a waypoint, and then a constraint is introduced to split the planning horizon at this waypoint so that planning can continue to the next waypoint or goal point.
Fig.~\ref{fig:wmpc_illustration} illustrates the proposed planning approach, which is described in Section~\ref{sec:rhtp}.

After discussing related work in Section~\ref{sec:related_work}, the mathematical model is introduced in Section~\ref{sec:math_model}.
%The core contribution of the waypoint MPC (WMPC) approach is described in Section~\ref{sec:rhtp}.
Subsequently, the proposed waypoint MPC (wMPC) is described in Section~\ref{sec:rhtp}.
The wMPC algorithm is compared to (global) RRT-type planners in a simulated environment in Section V-A to demonstrate that the trajectory duration and path length are comparable despite the local nature of the MPC.
A \kuka robot is used to experimentally demonstrate the online replanning capabilities of the proposed approach in a pick-and-place scenario. 
%The wMPC algorithm is compared to RRT-type planners in a simulated environment in Section~\ref{sec:sequential_manipulation} moving through several waypoints to demonstrate trajectory duration and path length competitiveness.
%In a dynamic pick-and-place task, the online replanning capabilities are demonstrated on a \kuka robot.
Section~\ref{sec:conclusion} concludes the paper and provides an outlook on future work.

%% file: sections/related_work.tex
%General trajectory-optimization / waypoint:
%TrajOpt: \cite{Schulman2014}
%CHOMP: \cite{Zucker2013}
%Gusto: \cite{BCBP2019}
%CIAO*: \cite{Schoels2020a}\cite{Schoels2020}
%
%Sequence-of-constraints mpc (most relevant): \cite{Toussaint2022} 
%Space-time functional gradient: \cite{Byravan2014}
%CALVIN (Task Benchmark): \cite{Mees2022}
%Logic-geometric programming: \cite{Toussaint2015}
%VP-STO: \cite{Jankowski2023}
%Path-following: \cite{Romero2022}, contouring \cite{Liniger2015} \cite{Lam2010}
%Time-optimal quadrotor waypoint: \cite{Foehn2021}

\subsection{Trajectory Optimization}

Classical trajectory optimization, e.g.,~\cite{Schulman2014, RZBS2009, BCBP2019, Jankowski2023}, optimizes an entire trajectory from an initial configuration to a goal. 
Waypoints can be introduced by constraining points along the trajectory.
If the trajectory duration is fixed, the timing for the waypoints must also be fixed.
On the other hand, if the trajectory duration is free and the end time serves as an additional optimization variable, the trajectory optimization problem becomes challenging to solve.
An efficient algorithm for calculating time-optimal trajectories through waypoints offline for quadrotor flight is proposed in \cite{Foehn2021}.

\subsection{MPC Through Waypoints}

Recent years have shown extensive interest in extending trajectory optimization to online planning using MPC. 
This includes gradient-based methods~\cite{GhazaeiArdakani2019, Kraemer2020, Schoels2020a} and sampling-based methods~\cite{Williams2016, Bhardwaj2022}.
However, these works do not explicitly consider the problem of going through desired waypoints. Therefore, several point-to-point motions must be planned for each waypoint, which implies either stopping or specifying a desired velocity at the waypoint in advance.

In contrast, an approach based on model-predictive contouring control for time-optimal quadrotor flight with waypoints was proposed in \cite{Romero2022}, where the waypoint timing is not predefined.
This approach relies on a pre-computed reference path through the waypoints. The MPC algorithm then tracks the path, allowing more significant deviations from the path between the waypoints to obtain an approximately time-optimal trajectory.
The reference path serves as a progress measure through the waypoints. 
However, it introduces additional complexity, which the task does not require since only passing the waypoints is necessary.
Furthermore, the authors do not investigate obstacle avoidance or dynamic replanning with changing waypoints.

The sequence-of-constraints MPC proposed in \cite{Toussaint2022} splits a task-and-motion-planning (TAMP) problem into three steps. 
First, the waypoints are obtained from planning a task. 
Second, the timing of the waypoints is optimized, resulting in a reference trajectory. 
In the third step, the reference trajectory is tracked with MPC to compute collision-free trajectories over a short planning horizon.
Similar to \cite{Romero2022}, a global reference is required to consider waypoints in the MPC.

In contrast to \cite{Toussaint2022}, the proposed approach does not compute a reference trajectory through all waypoints to determine their timing. 
Instead, the presented MPC formulation uses a cost-to-go towards the waypoints. 
It establishes a constraint for a specific timing of the waypoint as soon as the waypoint appears in the optimization horizon of the planner. 
Hence, the proposed approach does not need to compute a reference trajectory for the tracking MPC, which reduces the computational complexity and avoids problems with potentially infeasible reference trajectories.
%In particular, tracking a reference trajectory that is not collision-free can lead to additional local minima. 
%In particular, additional local minima caused by reference trajectories that are not collision-free are avoided.

In summary, the scientific contributions of this paper are three-fold: 
\begin{itemize}
    %\item The wMPC algorithm is proposed, a waypoint model predictive trajectory optimization algorithm that enables passing waypoints for sequential manipulation in a receding horizon manner while considering
    %dynamic constraints and obstacle avoidance for online replanning tasks without a global reference path or trajectory.
    \item The proposed wMPC algorithm enables model-predictive trajectory optimization through waypoints with a receding horizon for fast online replanning without a global reference.
    %\item The simulation results have shown that our wMPC achieves comparable results to global planners, i.e., RRT*, RRTConnect, and T-RRT, however, in an online fashion. 
    \item The simulation results show that our wMPC successfully traverses waypoints, and the planned trajectories result in similar durations and path lengths compared to RRT*, RRTConnect, and T-RRT in an online fashion.
    \item  The feasibility of the proposed wMPC is demonstrated experimentally in the online replanning application of a dynamic pick-and-place scenario for the \kuka robot. 
\end{itemize}

%Table~\ref{tab:comparison} summarizes the comparison to popular state-of-the-art algorithms. Note that time-optimal planning considers reasonable approximations, too.

%\begin{table}
%  \centering
%  \caption{\label{tab:comparison} Comparison with other state-of-the-art algorithms}
%  \scriptsize
%  \begin{tabular}{c | c | c | c | c | c}
%    \hline
%    Algorithm & waypoints & min. jerk & reactive & constraints & horizon \\
%    \hline\hline
%    RRT*~\cite{Karaman2011} & yes & no & no & yes & full \\
%    TrajOpt~\cite{Schulman2014} & fixed timing & yes & no & yes & full \\
%    CHOMP~\cite{RZBS2009} & no & yes & no & yes & full \\
%    VP-STO~\cite{Jankowski2023} & yes & no & yes & soft & full \\
%    CPC~\cite{Foehn2021} & yes & yes & yes & yes & full \\
%    STORM~\cite{Bhardwaj2022} & no & yes & yes & soft & receding \\
%    MPCC~\cite{Romero2022} & yes & yes & yes & yes & receding \\
%    SoC~\cite{Toussaint2022} & yes & yes & yes & yes & receding \\
%    ours & yes & yes & yes &  yes & receding \\
%    \hline
%  \end{tabular}
%\end{table}

%% file: sections/model.tex
The generalized coordinates $\vect{q} \in \mathbb{R}^m$ define the robot's configuration.
A double integrator model can be used, assuming that a suitable inverse dynamics control law, e.g.,~\cite{Ott2008}, compensates for the nonlinear dynamics of the robot manipulator.
For additional smoothness, however, a triple integrator model is used.
The state vector is defined as $\vect{x}^\mathrm{T} = [\vect{q}^\mathrm{T}, \dot{\vect{q}}^\mathrm{T}, \ddot{\vect{q}}^\mathrm{T}]$ with the input $\vect{u} = \dddot{\vect{q}}$.
%\begin{align}
%  \dot{\vect{x}} = \vect{A} \vect{x} + \vect{B} \vect{u}, \quad \vect{x}(0) = \vect{x}_0\Comma\label{eqn:sys}
%\end{align}
%with
%\begin{align}
%  \vect{A} = \begin{bmatrix}
%                  0 & 1 & 0 \\
%                  0 & 0 & 1 \\
%                  0 & 0 & 0    
%              \end{bmatrix}\otimes \vect{I}_{m}\Comma \quad
%  \vect{B} =  \begin{bmatrix}
%                  0 \\
%                  0 \\
%                  1
%              \end{bmatrix}\otimes \vect{I}_{m}\label{eqn:sys_mat}\Comma
%\end{align}
Assuming piecewise-linear inputs $\vect{u}_k$ with the sampling time $h$ leads to the first-order-hold discrete-time state-space formulation
\begin{align}
  \vect{x}_{k + 1} = \vect{\Phi} \vect{x}_k + \vect{\Gamma}_1 \vect{u}_k  + \vect{\Gamma}_2 \vect{u}_{k + 1}\label{eqn:sys_d}\Comma
\end{align}
where
\begin{align}
  \vect{\Phi} &= \begin{bmatrix}
                  1 & h & \frac{h^2}{2} \\
                  0 & 1 & h \\
                  0 & 0 & 1    
                \end{bmatrix}\otimes \vect{I}_{m}\Comma\quad\nonumber \\
  \vect{\Gamma}_1 &= \begin{bmatrix}
                  \frac{h^3}{8} \\
                  \frac{h^2}{3} \\
                  \frac{h}{2}
               \end{bmatrix} \otimes \vect{I}_{m} \Comma\quad
  \vect{\Gamma}_2 = \begin{bmatrix}
                  \frac{h^3}{24} \\
                  \frac{h^2}{6} \\
                  \frac{h}{2}
               \end{bmatrix} \otimes \vect{I}_{m}\FullStop\label{eqn:sys_d_mat}
\end{align} 
The symbol $\otimes$ denotes the Kronecker product, and $\vect{I}_m$ is the identity matrix of size $m$.

%The planner receives the waypoints and the goal in Cartesian space as a position vector and a quaternion.
%The analytic inverse kinematics solution from~\cite{Shimizu2008} for the \kuka robot used in this work calculates the corresponding joint configurations.

%% file: sections/rhtp.tex
This section presents the wMPC algorithm for trajectory optimization with waypoints over a receding horizon.
For formulating the optimization problem, a waypoint $\vect{q}_{\mathrm{w}}$ and a goal point $\vect{q}_{\mathrm{g}}$ in the joint space are considered.
The planner must pass the waypoint and finally stop at the goal point.
The MPC horizon length is initially set to its maximum $N_{\mathrm{max}}$ until the waypoint $\vect{q}_{\mathrm{w}}$ is reachable. 
Then, the horizon is split into two parts at the time index $N_{\mathrm{s}}$, where $(N_{\mathrm{s}} - 1)h$ refers to the time for reaching the waypoint $\vect{q}_{\mathrm{w}}$ and the remaining time from $N_{\mathrm{s}}h$ to $(N_{\mathrm{max}} - 1)h$ serves for planning towards the goal point $\vect{q}_{\mathrm{g}}$. 
The actual horizon length $N$ is then successively reduced when the goal point $\vect{q}_{\mathrm{g}}$ appears within the horizon. 
Section~\ref{sec:planning_algorithm} discusses in more detail how to split the maximum horizon $N_{\mathrm{max}}$ and how to calculate $N_{\mathrm{s}}$ and the reduction of the horizon length.
%To this end, the MPC horizon is split into two parts at the index $N_{\mathrm{s}}$. 
%At the start, $N_{\mathrm{s}}$ covers the entire horizon length $N_{\mathrm{max}}$ until the waypoint $\vect{q}_{\mathrm{w}}$ is reachable. It will then be reduced such that the remaining horizon until the index $N$ can continue planning towards the goal point $\vect{q}_{\mathrm{g}}$, which is again shrinking as soon as the goal point is reachable.
%The split of the full horizon $N_{\mathrm{max}}$ and how to calculate $N_{\mathrm{s}}$ and $N$ will be discussed in Section~\ref{sec:planning_algorithm}.

\subsection{Optimization Problem}
\label{sec:opt_problem}

The computation of the optimal trajectory for the system state $\vect{x}_{0 | n}, \dots, \vect{x}_{N - 1 | n}$ and the system input $\vect{u}_{0|n}, \dots, \vect{u}_{N - 1 | n}$ for the MPC iteration $n$ is formulated as the discrete-time optimization problem
\begin{subequations}
  \label{eqn:jerk_opt}
  \begin{alignat}{2}
    &\min_{\substack{\vect{x}_{0 | n}, \dots, \vect{x}_{N - 1 | n}, \\\vect{u}_{0 | n}, \dots, \vect{u}_{N - 1 | n}}} &&\sum_{k = 0}^{N_{\mathrm{s}} - 1}  w_1 l_1(\vect{x}_{k|n}) + \sum_{k = N_{\mathrm{s}}}^{N - 1}  w_2 l_2(\vect{x}_{k|n}) \nonumber \\ &\quad && + \sum_{k = 0}^{N - 1} \dist{\vect{u}_{k|n}}_2^2 + w_3 l_{\mathrm{col}}(\vect{x}_{k|n}) \label{eqn:opt_cost}\\
    &\quad \text{s.t.} && \vect{x}_{k + 1 | n} = \vect{\Phi} \vect{x}_{k | n } + \vect{\Gamma}_1 \vect{u}_{k | n} + \vect{\Gamma}_2 \vect{u}_{k + 1 | n},\nonumber \\ &\quad && k = 0, \dots, N - 2 \label{eqn:opt_dyn}\\
    & \quad && \vect{x}_{0 | n} = \vect{x}_{1 | n - 1},\quad \vect{u}_{0 | n} = \vect{u}_{1 | n - 1} \label{eqn:init_cond_1}\\
    %& \quad && \vect{u}_{0 | n} = \vect{u}_{1 | n - 1} \label{eqn:init_cond_2}\\
    & \quad && \vect{x}_{N - 1 | n} = \vect{\Phi} \vect{x}_{N - 1 | n},\quad \vect{u}_{N - 1 | n} = \vect{0}\label{eqn:steady_state_x}\\
    %& \quad && \vect{u}_{N - 1 | n} = \vect{0} \label{eqn:steady_state_u}\\
    &                  \quad && \underline{\vect{x}} \le \vect{x}_{k | n} \le \overline{\vect{x}},\quad \underline{\vect{u}} \le \vect{u}_{k | n} \le \overline{\vect{u}} \label{eqn:x_limit} \\
    %&             \quad &&\underline{\vect{u}} \le \vect{u}_{k | n} \le \overline{\vect{u}} \label{eqn:u_limit} \\
    & \quad && \vect{q}_{N_{\mathrm{s}} - 1} \in \mathcal{Q}_{\mathrm{w}},\quad \vect{q}_{N - 1} \in \mathcal{Q}_{\mathrm{g}} \label{eqn:end_point_way} %\\
    %& \quad && \vect{q}_{N - 1} \in \mathcal{Q}_{\mathrm{d}} \label{eqn:end_point_goal}
  \end{alignat}
\end{subequations}
where (\ref{eqn:opt_dyn}) ensures the trajectory adheres to the system dynamics.
The initial states are given by (\ref{eqn:init_cond_1}) for the system state and input, where $\vect{x}_{1 | n - 1}$ and $\vect{u}_{1 | n - 1}$ result from the previous MPC iteration.
In order to ensure that the final state in the horizon is a steady state, (\ref{eqn:steady_state_x}) is required, c.f.~\cite{Schoels2020a}.
The advantage of always ending in a steady state is that each optimized trajectory is valid and safe, resulting in an anytime property of the wMPC algorithm for static environments.
For the trajectory to be executable on the robot, boundary constraints on the states and inputs (\ref{eqn:x_limit}) must be fulfilled, with the lower limits $\underline{\vect{x}}$, $\underline{\vect{u}}$ and the upper limits $\overline{\vect{x}}$, $\overline{\vect{u}}$.
%, and the inputs (\ref{eqn:u_limit}), with the lower limit $\underline{\vect{u}}$ and the upper limit $\overline{\vect{u}}$ are applied.
The final point $\vect{q}_{N_{\mathrm{s}} - 1}$ in the first part of the horizon up to $N_{\mathrm{s}} - 1$ must be in the set $\mathcal{Q}_{\mathrm{w}}$ such that the waypoint is passed and the final point $\vect{q}_{N - 1}$ of the overall horizon in the set $\mathcal{Q}_{\mathrm{g}}$, which is ensured by (\ref{eqn:end_point_way}).
Depending on the reachability of the waypoint $\vect{q}_{\mathrm{w}}$ and the goal point $\vect{q}_{\mathrm{g}}$, $N_\mathrm{s}$ and $N$ will be reduced, as discussed in Section~\ref{sec:planning_algorithm}.
The shrinking horizons ensure that only the minimum amount of required samples is used for planning, which avoids oscillations towards the end of the trajectory.

Two cases must be distinguished to determine the terminal constraint sets $\mathcal{Q}_{\mathrm{w}}$ and $\mathcal{Q}_{\mathrm{g}}$.
First, if the waypoint or the goal point is not reachable within the horizons $N_{\mathrm{s}} - 1$ or $N - 1$, respectively, the terminal constraint sets $\mathcal{Q}_{\mathrm{w}}$ and $\mathcal{Q}_{\mathrm{g}}$ are only restricted by the joint limits of the robot.
Otherwise, the sets $\mathcal{Q}_{\mathrm{w}}$ and $\mathcal{Q}_{\mathrm{g}}$ are defined by a tolerance band around the waypoint $\vect{q}_\mathrm{w}$ and the goal point $\vect{q}_{\mathrm{g}}$ for each component $i = 0, \dots, m-1$, with the tolerance distance $\varepsilon > 0$, see (\ref{eqn:qw}) and (\ref{eqn:qd}). 
Thus, the sets $\mathcal{Q}_{\mathrm{w}}$ and $\mathcal{Q}_{\mathrm{g}}$ are defined as
\begin{equation}
  \label{eqn:qw}
  \mathcal{Q}_{\mathrm{w}} = \begin{cases}
    \begin{aligned} &\{ \vect{q}~|~|q_i - q_{i, \mathrm{w}}| \le \varepsilon,\\&\phantom{\{ \vect{q}~|~}i = 0, \dots, m - 1 \}, \end{aligned} &\quad N_{\mathrm{s}} < N - 1 \\
    \{ \vect{q}~|~\underline{\vect{q}} \le \vect{q} \le \overline{\vect{q}} \}, &\quad \text{otherwise}  \Comma\\
  \end{cases}
\end{equation}
and
\begin{equation}
  \label{eqn:qd}
  \mathcal{Q}_{\mathrm{g}} = \begin{cases}
    \begin{aligned} &\{ \vect{q}~|~|q_i - q_{i, \mathrm{g}}| \le \varepsilon,\\&\phantom{\{ \vect{q}~|~}i = 0, \dots, m - 1 \}, \end{aligned} &\quad N - 1 < N_{\mathrm{max}} \\
    \{ \vect{q}~|~\underline{\vect{q}} \le \vect{q} \le \overline{\vect{q}} \}, &\quad \text{otherwise} \FullStop
  \end{cases}
\end{equation}

\begin{remark}
    The change in the terminal constraint sets $\mathcal{Q}_{\mathrm{w}}$ and $\mathcal{Q}_{\mathrm{g}}$ when $\vect{q}_{\mathrm{w}}$ or $\vect{q}_{\mathrm{g}}$ become reachable does not impact the recursive feasibility of the optimization problem.
    When the environment is static, the reachability in a previous iteration implies reachability in the next iteration.
    If the environment changes, recursive feasibility is not ensured.
    However, in that case, $N_{\mathrm{s}}$ and $N$ are reset, and $\mathcal{Q}_{\mathrm{w}}$ and $\mathcal{Q}_{\mathrm{g}}$ contain the robot's reachable workspace again.
\end{remark}

%\todo{Introduce shrinking horizon}

The objective functions $l_1(\vect{x}_{k|n})$ and $l_2(\vect{x}_{k|n})$ with the weights $w_1, w_2 > 0$ in (\ref{eqn:opt_cost}) give a cost-to-go towards the waypoint and the goal point, respectively.
%The cost-to-go should be chosen in according to the tolerance interval defined in (\ref{eqn:qw}) and (\ref{eqn:qd}).
%Here, a smooth approximation of the 1-norm is used, resulting in
The cost-to-go is chosen as a smooth approximation of the 1-norm, resulting in
\begin{align}
  \label{eqn:l1}
  l_1(\vect{x}_{k|n}) &= \sum_{i = 0}^{m - 1} \sqrt{(q_{k, i|n} - q_{\mathrm{w}, i})^2 + \gamma^2} - \gamma \Comma
\end{align}
and
\begin{align}
  \label{eqn:l2}
  l_2(\vect{x}_{k|n}) &= \sum_{i = 0}^{m - 1} \sqrt{(q_{k, i|n} - q_{\mathrm{g}, i})^2 + \gamma^2} - \gamma \Comma
\end{align}
with a parameter $\gamma > 0$.
For smaller $\gamma$, the approximation is more accurate. 
Convergence difficulties can occur if $\gamma$ is too small because the gradient increases close to the waypoint and goal point, respectively.
%\begin{remark}
%  Note that the cost-to-go terms are chosen to only depend on the joint position $\vect{q}_{k|n}$ but not on the joint velocity $\dot{\vect{q}}_{k|n}$ or the joint acceleration $\ddot{\vect{q}}_{k|n}$ because the planner is forced to stop at the goal due to the constraints (\ref{eqn:steady_state_x}). However, an additional penalty for regularization could be realized using the objective function.
%\end{remark}

\begin{remark}
  Choosing a 1-norm cost function for (\ref{eqn:l1}) and (\ref{eqn:l2}) has additional advantages in terms of the qualitative properties of the planned trajectories through waypoints. 
  However, it may entail numerical issues due to the discontinuity of the gradient at the waypoint and goal point.
  %A 2-norm typically leads to a more prolonged phase where the trajectory slows down towards a goal due to the quadratic cost reduction. 
  %Section~\ref{sec:waypoint_transversal} compares 1-norm and 2-norm cost terms in the planning algorithm.
\end{remark}

The objective function $l_{\mathrm{col}}(\vect{x}_{k|n})$ in (\ref{eqn:opt_cost}) is a collision avoidance term with the weight $w_3 > 0$.
Calculating the distances to the obstacles is outside the scope of this paper. 
It is assumed that a signed distance $d_{i, j}(\vect{q}_{k|n})$ between each collision object $\mathcal{O}_i$, $i = 0, \dots, N_{O} - 1$ and each part of the collision model of the robot (including the gripper) $\mathcal{R}_j$, $j = 0, \dots, N_{R} - 1$ is available.
The signed distance is easily calculated for simple geometries, like spheres and capsules. For more complex geometries, algorithms exist in the literature, e.g.,~\cite{Cameron1997}.
Similar to~\cite{Vu2020}, a smooth approximation of the maximum function is employed, resulting in the collision cost term
\begin{align}
  \varphi_{i, j}(\vect{q}_{k|n}) = \frac{1}{\alpha} \log\left(1 + \exp(-\alpha(d_{i, j}(\vect{q}_{k|n}) + \beta))\right) \Comma
\end{align}
with the parameters $\alpha > 0$ describing the steepness of the approximation and $\beta > 0$ shifts the curve such that $\varphi_{i, j}(\vect{q}_{k|n}) > 0$ only if the robot is close to contact.
The overall collision objective function $l_{\mathrm{col}}(\vect{x}_{k|n})$ is then
\begin{align}
  l_{\mathrm{col}}(\vect{x}_{k|n}) = \sum_{i = 0}^{N_O - 1} \sum_{j = 0}^{N_R - 1} \varphi_{i, j}(\vect{q}_{k|n}) \FullStop
\end{align}

The weights $w_1$ and $w_2$ in (\ref{eqn:opt_cost}) are chosen indirectly proportional to the distances between the starting point $\vect{q}_{\mathrm{init}}$ and the waypoint $\vect{q}_{\mathrm{w}}$ and between the waypoint $\vect{q}_{\mathrm{w}}$ and the goal point $\vect{q}_{\mathrm{g}}$, respectively.
This results in
\begin{align}
  %\label{eqn:weights}
  w_1 &= \frac{\sigma}{\max(\dist{\vect{q}_{\mathrm{w}} - \vect{q}_{\mathrm{init}}}_2, d_{\mathrm{min}})}\label{eqn:w1} \\
  w_2 &= \frac{\sigma}{\max(\dist{\vect{q}_{\mathrm{g}} - \vect{q}_{\mathrm{w}}}_2, d_{\mathrm{min}})}\label{eqn:w2} \Comma
\end{align}
where $\sigma > 0$ is a scaling factor, and $d_{\mathrm{min}} > 0$ prevents division by zero.
By choosing the weights according to (\ref{eqn:w1}) and (\ref{eqn:w2}), the planner computes trajectories that take less time for shorter segments, i.e., the weights for the cost-to-go become larger for shorter segments.
Hence, similar distances require similar time, making the trajectory's velocity profile consistent throughout the planned segments.
When $\sigma$ is increased, the resulting trajectories are more aggressive, resulting in higher velocity.
The planner can achieve approximately time-optimal behavior for large $\sigma$ and $N$, c.f.~\cite{Verschueren2017}.
The weight for the collision avoidance $w_3$ has to be larger than $w_1$ and $w_2$ to ensure collision avoidance since no constraint for collision avoidance is present in the planner.

%\begin{figure}[!tb]
% \removelatexerror
%  \begin{algorithm}[H]
%    \caption{\label{alg:waypoint_point_cutoff} Calculation of $N_{\mathrm{s}}$}
%    \DontPrintSemicolon
%    \SetKwInOut{Input}{Input}\SetKwInOut{Output}{Output}
%    \Input{$N_{\mathrm{s}}, \vect{Q} = [\vect{q}_0, \dots, \vect{q}_{N - 1}]^\mathrm{T}, \vect{q}_{\mathrm{w}}$}
%    \Output{$N_{\mathrm{s}}$}
%    
%    \If{$N_}
%
%  \end{algorithm}
%\end{figure}

%\begin{figure}[!tb]
% \removelatexerror
%  \begin{algorithm}[H]
%    \caption{\label{alg:horizon_length} Calculation of $N$}
%    \DontPrintSemicolon
%    \SetKwInOut{Input}{Input}\SetKwInOut{Output}{Output}
%    \Input{$N_{\mathrm{max}}, \vect{q}_{\mathrm{w}}, \vect{q}_{\mathrm{d}}$}
%    \Output{$N$}
%    $i \leftarrow 0$\;
%    $d \leftarrow \dist{\vect{q}_{\mathrm{w}} - \vect{q}_{\mathrm{d}}}_2$\;
%    $d_{\mathrm{prev}} \leftarrow \dist{\vect{q}_{\mathrm{w}} - \vect{q}_{\mathrm{d}}}_2$\;
%    $N \leftarrow N_{\mathrm{max}}$\;
%
%    \If{$N_{\mathrm{s}} < N - 2$}
%    {
%      \eIf{$N$ = $N_{\mathrm{max}}$}
%      {
%        \While{$i < N - 1 \land d > \varepsilon \land d_{\mathrm{prev}} \ge d$}
%        {
%          $d \leftarrow \dist{\vect{q}_i - \vect{q}_{\mathrm{d}}}_2$\;
%          $i \leftarrow i + 1$\;
%        }
%        $N \leftarrow \max(i - 1, 2)$
%      }
%      {
%        $N \leftarrow \max(N - 1, 2)$\;
%      }
%    }
%  \end{algorithm}
%\end{figure}

\subsection{Planning Algorithm}
\label{sec:planning_algorithm}

Algorithm~\ref{alg:planning} plans from the current robot state $\vect{x}_{1|n - 1}$ to a Cartesian goal pose, described by the homogeneous transformation $\vect{T}_{\mathrm{g}}$, through a waypoint described by $\vect{T}_{\mathrm{w}}$.
If a new goal arrives, the horizon lengths $N_{\mathrm{s}}$ and $N$ are set to the maximum horizon length $N_{\mathrm{max}}$, and $\vect{q}_{\mathrm{w}}$ and $\vect{q}_{\mathrm{g}}$ are calculated by an inverse kinematics algorithm.
Then, the state and input trajectories are initialized using the solution of a previous MPC iteration if available, and the weights $w_1$ and $w_2$ are computed according to (\ref{eqn:w1}) and (\ref{eqn:w2}). 
Lines 1 - 10 of Algorithm~\ref{alg:planning} show this procedure.

In lines 11 - 13, the planner examines whether the waypoint $\vect{q}_{\mathrm{w}}$ is reachable within the maximum horizon length $N_{\mathrm{max}}$ using Algorithm~\ref{alg:check_goal_distance}.
This algorithm checks whether the components $q_{\mathrm{g}, j}$, $j = 0, \dots, m - 1$, of a goal point $\vect{q}_{\mathrm{g}}$ (or a waypoint $\vect{q}_{\mathrm{w}}$) can be reached within the tolerance band $\varepsilon$ in the interval $[N_{\mathrm{start}}, \dots, N_{\mathrm{stop} - 1}]$, see lines 5 - 11.
%This algorithm iterates over the planned trajectory indices from $N_{\mathrm{start}}$ to $N_{\mathrm{stop}} - 1$ in lines 2 - 17 and checks for each joint $q_{i, j}$, $i = N_{\mathrm{start}}, \dots, N_{\mathrm{stop}} - 1$, $j = 0, \dots, m - 1$ if the goal is reachable within the tolerance $\varepsilon$, see lines 5 - 11.
For this purpose, the boolean array $reached$ in line 1 of Algorithm~\ref{alg:check_goal_distance} keeps track of which joints can reach their goal.
Even if not all components $j = 0, \dots, m - 1$ at a time instant $i \in [N_{\mathrm{start}}, \dots, N_{\mathrm{stop} - 1}]$ satisfy the condition $|q_{i,j} - q_{\mathrm{g}, j}| < \varepsilon$, the goal point $\vect{q}_{\mathrm{g}}$ is reachable if $\sign(q_{i, j} - q_{\mathrm{g},j}) \neq \sign(q_{i - 1, j} - q_{\mathrm{g}, j})$ is fulfilled. 
Fig. 2 illustrates such a case for $m = 2$, where the goal point $\vect{q}_{\mathrm{g}}$ is reachable within the tolerance band $\varepsilon$ although $|q_{i - 1,0} - q_{\mathrm{g}, 0}| > \varepsilon$ and $|q_{i, 0} - q_{\mathrm{g}, 0}| > \varepsilon$ since the connecting line goes through the tolerance band, which is indicated by the change in sign of $q_{i - 1, 0} - q_{\mathrm{g}, 0}$ and $q_{i, 0} - q_{\mathrm{g}, 0}$.

%If the goal is not reached within the tolerance $\varepsilon$ for a particular joint $q_{i, j}$, the joint may have already passed the goal in this time step but not within the tolerance.
%In this case, the goal of the joint can still be considered reachable and hence added to the $reached$ array. 
%Fig.~\ref{fig:reachability_sketch} illustrates this case in a 2D example. 
%The goal is reachable within the tolerance only for the second joint in step $i - 1$. 
%In the next step, $i$, the first joint is still not within the tolerance but has passed the goal between the trajectory points.
\begin{figure}[ht]
  \centering
  \medskip
  \includegraphics{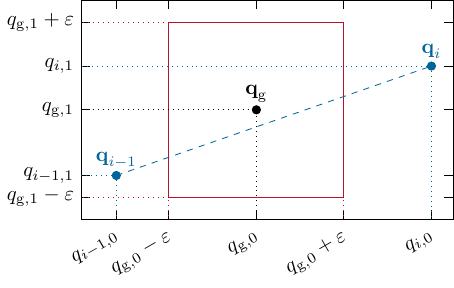}
  \caption{\label{fig:reachability_sketch} This figure illustrates when a goal point $\vect{q}_{\mathrm{g}}$ counts as reachable within the horizon for $m = 2$. First, if all components $j = 0, \dots, m - 1$ of a point $\vect{q}_{i}$ are within the tolerance band $\varepsilon$, then the goal is reachable. In this example, this is only the case for the second component $q_{i - 1, 1}$ and $q_{i, 1}$. However, it is evident for the first component that the connecting line between $q_{i - 1, 0}$ and $q_{i, 0}$ goes through the tolerance band.}
\end{figure}

If the waypoint $\vect{q}_{\mathrm{w}}$ is reachable at a time instant $i < N_{\mathrm{max}}$, then $N_{\mathrm{s}} = i - 1$; see lines 11 - 14 of Algorithm~\ref{alg:planning}.
Analogously, if the goal point $\vect{q}_{\mathrm{g}}$ can be reached at a time instant $i$ within the maximum horizon length $N_{\mathrm{max}}$, the actual horizon length $N$ is chosen as $N = i - 1$; see lines 15 - 20 of Algorithm~\ref{alg:planning}.
\begin{remark}
  Note that when the goal is reachable for the first time, i.e., line 16 returns a value smaller than $N_{\mathrm{max}}$, an appropriate minimum horizon length must be chosen.
  A minimum length of at least three steps (dead-beat behavior) is necessary to drive the dynamics~(\ref{eqn:opt_dyn}) from an initial condition~(\ref{eqn:init_cond_1}) to the goal~(\ref{eqn:end_point_way}) without state and input constraints.
  For the practical implementation, this minimum horizon length was increased to 5.
  %Note that when the goal is reachable for the first time, i.e., line 16 returns a value smaller than $N_{\mathrm{max}}$, the minimum horizon length is chosen as $5$, ensured in line 17, even if the planner could reach $\vect{q}_{\mathrm{g}}$ earlier.
  %A minimum length of at least three steps is necessary to drive the dynamics~(\ref{eqn:opt_dyn}) from an initial condition~(\ref{eqn:init_cond_1}) to the goal~(\ref{eqn:end_point_way}).
\end{remark}

The sets $\mathcal{Q}_{\mathrm{w}}$ and $\mathcal{Q}_{\mathrm{g}}$ are determined in line 22 of Algorithm~\ref{alg:planning} according to (\ref{eqn:qw}) and (\ref{eqn:qd}), respectively.
Then, the optimization problem (\ref{eqn:jerk_opt}) is solved to obtain the optimal trajectory planning result, and the first step of the trajectories $\vect{x}_{0|n}$ and $\vect{u}_{0|n}$ is sent to the controller and executed on the robot.

In future iterations of the same waypoint and goal point, the planner repeats the reachability checks if the waypoint or goal point was not reachable before.
Otherwise, the horizon lengths are reduced by one in each iteration.
The minimum horizon length towards the waypoint $N_\mathrm{s}$ is zero; see line 14 in Algorithm~\ref{alg:planning}.
In contrast, the minimum horizon length for the goal point $N$ is two, see line 19 of Algorithm~\ref{alg:planning}, because the first step in the optimization (\ref{eqn:jerk_opt}) is already constrained to the initial value of the previous MPC iteration in (\ref{eqn:init_cond_1}).

\begin{figure}[!t]
 \removelatexerror
 \medskip
  \begin{algorithm}[H]
    \caption{\label{alg:planning} wMPC Motion Planning Algorithm}
    \DontPrintSemicolon
    \SetKwData{newG}{new\_goal}
    \SetKwFunction{inverseKinematics}{inverseKinematics}
    \SetKwFunction{computeWeights}{computeWeights}
    \SetKwFunction{initializeTrajectory}{initializeTrajectory}
    \SetKwFunction{calculateNs}{calculateNs}
    \SetKwFunction{calculateN}{calculateN}
    \SetKwFunction{WMPCOPT}{WMPC-OPT}
    \SetKwFunction{checkReach}{checkGoalReachability}
    \SetKwInOut{Input}{Input}\SetKwInOut{Output}{Output}
    \Input{$\vect{T}_{\mathrm{w}}$, $\vect{T}_{\mathrm{g}}$, \newG, $\vect{x}_{1|n-1}$, $\vect{u}_{1|n-1}$}
    \Output{$\vect{x}_{0 | n}, \dots, \vect{x}_{N - 1 | n}, \vect{u}_{0 | n}, \dots, \vect{u}_{N - 1 | n}$}

    \If{\newG}
    {
      $N_{\mathrm{s}} = N_{\mathrm{max}}$\;
      $N = N_{\mathrm{max}}$\;
      $\vect{q}_{\mathrm{w}} \leftarrow~$\inverseKinematics{$\vect{T}_{\mathrm{w}}$}\;
      $\vect{q}_{\mathrm{g}} \leftarrow~$\inverseKinematics{$\vect{T}_{\mathrm{g}}$}\;
      $\vect{x}_{0 | n}, \dots, \vect{x}_{N - 1 | n}, \vect{u}_{0 | n}, \dots, \vect{u}_{N - 1 | n} \leftarrow~$\;
      \pushline\initializeTrajectory{}\;
      \popline
      $\vect{q}_{\mathrm{init}} \leftarrow \vect{q}_{0|n}$\;
      \computeWeights{}\;
    }

    \eIf{$N_{\mathrm{s}} = N_{\mathrm{max}}$}
    {
      $N_{\mathrm{s}} \leftarrow $\checkReach{
      \pushline$0, N_{\mathrm{s}}, \vect{q}_{\mathrm{w}}, [\vect{q}_{0|n}, \dots, \vect{q}_{N_{\mathrm{s}} - 1|n}]^\mathrm{T}$}\;
      \popline
    }
    {
      $N_{\mathrm{s}} \leftarrow \max(N_{\mathrm{s}} - 1, 0)$\;
      \eIf{$N = N_{\mathrm{max}}$}
      {
        $N \leftarrow $\checkReach{
          \pushline$N_{\mathrm{s}}, N, \vect{q}_{\mathrm{g}}, [\vect{q}_{N_s|n}, \dots, \vect{q}_{N - 1|n}]^\mathrm{T}$}\;
          $N \leftarrow \max(N, 5)$\;
      }
      {
        $N \leftarrow \max(N - 1, 2)$
      }
    }

    Compute $\mathcal{Q}_{\mathrm{w}}$ and $\mathcal{Q}_{\mathrm{g}}$ using (\ref{eqn:qw}) and (\ref{eqn:qd})

    %\eIf{$N_{\mathrm{s}} < N_{\mathrm{max}}$}
    %{
    %  $\mathcal{Q}_{\mathrm{w}} \leftarrow \{ \vect{q}~|~|q_i - q_{i, \mathrm{w}}| \le \varepsilon,~i = 0, \dots, m - 1 \}$\;
    %}
    %{
    %  $\mathcal{Q}_{\mathrm{w}} \leftarrow \{ \vect{q}~|~\underline{\vect{q}} \le \vect{q} \le \overline{\vect{q}} \}$\;
    %}
    %\eIf{$N < N_{\mathrm{max}}$}
    %{
    %  $\mathcal{Q}_{\mathrm{d}} \leftarrow \{ \vect{q}~|~|q_i - q_{i, \mathrm{d}}| \le \varepsilon,~i = 0, \dots, m - 1 \}$\;
    %}
    %{
    %  $\mathcal{Q}_{\mathrm{d}} \leftarrow \{ \vect{q}~|~\underline{\vect{q}} \le \vect{q} \le \overline{\vect{q}} \}$\;
    %}
    $\vect{x}_{0 | n}, \dots, \vect{x}_{N - 1 | n}, \vect{u}_{0 | n}, \dots, \vect{u}_{N - 1 | n} \leftarrow~$
    \pushline solve optimization problem (\ref{eqn:jerk_opt})
    %\WMPCOPT{$N_{\mathrm{s}}, N, \vect{q}_{\mathrm{w}}, \vect{q}_{\mathrm{g}}, \vect{x}_{1|n - 1}, \vect{u}_{1|n-1}$}\;
    \popline
  \end{algorithm}
\end{figure}

\begin{figure}[!t]
 \removelatexerror
  \begin{algorithm}[H]
    \caption{\label{alg:check_goal_distance} Check Goal Reachability}
    \DontPrintSemicolon
    \SetKwData{reached}{reached}
    \SetKwFunction{zeros}{zeros}
    \SetKwInOut{Input}{Input}\SetKwInOut{Output}{Output}
    \Input{$N_{\mathrm{start}}, N_{\mathrm{stop}}, \vect{q}_{\mathrm{g}}, [\vect{q}_{N_{\mathrm{start}}}, \dots, \vect{q}_{N_{\mathrm{stop}}-1}]$}
    \Output{Index of the trajectory that reaches the goal}

    $\reached \leftarrow \zeros{m}$\;

    \For{$i \leftarrow N_{\mathrm{start}}$ \KwTo $N_{\mathrm{stop}} - 1$}
    {
      \For{$j \leftarrow 0$ \KwTo $m - 1$}
      {
        $d \leftarrow q_{i, j} - q_{\mathrm{g}, j}$\;
        \If{$|d| < \varepsilon$}
        {
          $\reached(j) \leftarrow 1$\;
        }
        \If{$i > 0$}
        {
          \If{$\sign(d) \neq \sign(q_{i - 1, j} - q_{\mathrm{g}, j})$}
          {
            $\reached(j) \leftarrow 1$\;
          }
        }
      }
      \If{all entries of \reached are 1}
      {
        \Return $i$
      }
    }
    \Return $i + 1$

  \end{algorithm}
\end{figure}

\subsection{Extension to Multiple Waypoints}
\label{sec:multiple_waypoints}

The presented wMPC algorithm can be readily extended to a sequence of waypoints $\mathcal{W} = \{\vect{q}_{\mathrm{w}, 0}, \vect{q}_{\mathrm{w}, 1}, \dots, \vect{q}_{\mathrm{w}, N_{\mathrm{way}} - 1}\}$.
There are two possibilities to achieve this.
On the one hand, the optimization problem~(\ref{eqn:jerk_opt}) can be extended to include several horizons instead of only two.
The main advantage of this approach is that several waypoints can be considered simultaneously during the optimization, which can be necessary if the waypoints lie close together.
However, this is not easy to implement because the number of waypoints is unknown in advance, and each waypoint adds computational complexity.
Therefore, on the other hand, only one waypoint and one goal point are considered in the optimization problem. 
The current waypoint and goal point are chosen according to which waypoints the robot has passed.
%Algorithm~\ref{alg:waypoint_sequence} summarizes the procedure called before every iteration of the WMPC planning algorithm.
%\begin{figure}[!tb]
% \removelatexerror
%  \begin{algorithm}[H]
%    \caption{\label{alg:waypoint_sequence} Waypoint Sequence Scheduling}
%    \DontPrintSemicolon
%    \SetKwInOut{Input}{Input}\SetKwInOut{Output}{Output}
%    \Input{$\mathcal{W}~=~\{\vect{q}_{\mathrm{w}, 0}, \vect{q}_{\mathrm{w}, 2}, \dots, \vect{q}_{N_{\mathrm{way}} - 1}\}, N_{\mathrm{way}}, \newline N_{\mathrm{s}}, c$}
%    \Output{$\vect{q}_{\mathrm{d}}, \vect{q}_{\mathrm{d}}$}
%
%    \If{$N_{\mathrm{s}} = 0 \land c < N_{\mathrm{way}} - 1$}
%    {
%      $\vect{q}_{\mathrm{w}} \leftarrow \vect{q}_{\mathrm{w}, c}$\;
%      $\vect{q}_{\mathrm{d}} \leftarrow \vect{q}_{\mathrm{w}, c + 1}$\;
%      $c = c + 1$\;
%    }
%  \end{algorithm}
%\end{figure}
A waypoint is considered as reached if $N_{\mathrm{s}}$ becomes zero.
In this case, the current goal $\vect{q}_{\mathrm{g}} = \vect{q}_{\mathrm{w}, c}$ is the new waypoint $\vect{q}_{\mathrm{w}} = \vect{q}_{\mathrm{w}, c}$, and the next waypoint in the sequence $\vect{q}_{\mathrm{w}, c + 1}$ is chosen as the new goal $\vect{q}_{\mathrm{g}} = \vect{q}_{\mathrm{w}, c + 1}$ for the wMPC planner. 

%% file: sections/results.tex
The presented algorithm is demonstrated for two scenarios on a \kuka robot with 7-DoF.
%First, the qualitative behavior of the algorithm is investigated for planning through several waypoints with different parameters for the velocity scaling factor $\sigma$, the maximum horizon length $N_{\mathrm{max}}$. Furthermore, using squared 2-norms in the cost functions $l_1(\vect{x}_{k|n})$ and $l_1(\vect{x}_{k|n})$ is compared to using a 1-norm approximation.
\begin{table}[t]
  \centering
  \medskip
  \caption{\label{tab:wmpc_params} Planning Algorithm Parameters}
  \begin{tabular}{c c c c c c c c c }
    \hline
    $h$ & $N_{\mathrm{max}}$ & $w_3$ & $\varepsilon$ & $\gamma$ & $\alpha$ & $\beta$ & $\sigma$ & $d_{\mathrm{min}}$ \\
    \hline\hline
    0.1 & 20 & 100 & 0.0005 & 0.1 & 1000 & 0.001 & 20 & 0.01 \\
    \hline
  \end{tabular}
\end{table}
\begin{table}[t]
  \caption{\label{tab:bounds} Upper and lower bounds for $\vect{q}$, $\qdot$, and $\ddot{\vect{q}}$.}
  \centering
  \begin{tabular}{c | c | c}
    \hline
    Symbol & Value & Unit \\
    \hline
    \hline
    $\overline{\vect{q}}$, $\underline{\vect{q}}$ & $\pm\frac{\pi}{180}\left[170, 120, 170, 120, 170, 120, 175\right]^\mathrm{T}$ &  \SI{}{\radian}\\
    %$\underline{\vect{q}}$ & $-\frac{\pi}{180}\left[170, 120, 170, 120, 170, 120, 175\right]^\mathrm{T}$ & \SI{}{\radian} \\
    $\overline{\dot{\vect{q}}}$, $\underline{\dot{\vect{q}}}$ & $\pm\frac{\pi}{180}\left[85, 85, 100, 75, 130, 135, 135\right]^\mathrm{T}$ & \SI{}{\radian\per\second}\\
    $\overline{\ddot{\vect{q}}}$, $\underline{\ddot{\vect{q}}}$ & $\pm\left[5, 5, 5, 5, 5, 5, 5\right]^\mathrm{T}$ & \SI{}{\radian\per\square\second}\\
    %$\underline{\dot{\vect{q}}}$ & $-\frac{\pi}{180}\left[85, 85, 100, 75, 130, 135, 135\right]^\mathrm{T}$ & \SI{}{\radian\per\second}\\
    \hline
  \end{tabular}
\end{table}
%\begin{figure*}[!ht]
%  \centering
%  \subfloat[]{\includegraphics[width = 0.49\columnwidth]{graphics/sim/start.png}} \hfil
%  \subfloat[]{\includegraphics[width = 0.49\columnwidth]{graphics/sim/push.png}} \hfil
%  \subfloat[]{\includegraphics[width = 0.49\columnwidth]{graphics/sim/pick2.png}} \hfil
%  \subfloat[]{\includegraphics[width = 0.49\columnwidth]{graphics/sim/place3.png}} \hfil
%  \caption{\label{fig:sim} Sequential manipulation task in MuJoCo~\cite{Todorov2012}. The robot starts from an initial configuration in (a) and then moves through a sequence of waypoints to slide open the cabinet door in (b). Afterwards, the robot must avoid the cylindrical obstacle while approaching and grasping the object in (c). Finally, the robot places the object into the cabinet in (d).}
%\end{figure*}
\begin{figure}[!t]
  \centering
  \subfloat[]{\includegraphics[width = 0.42\columnwidth]{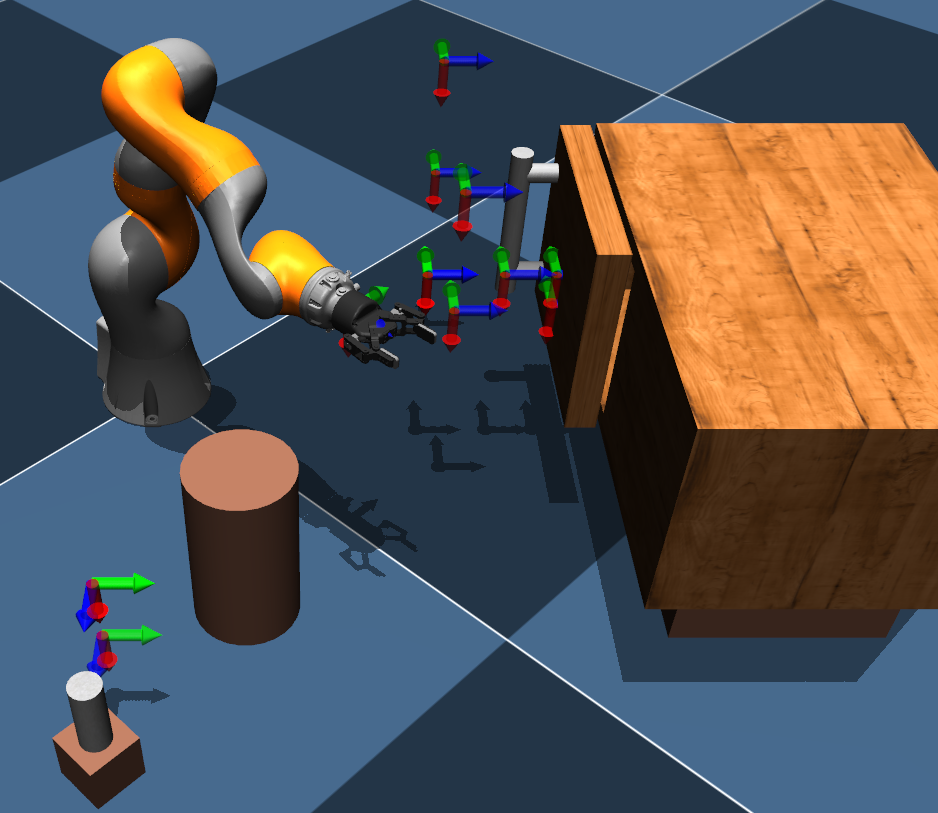}} \hfil
  \subfloat[]{\includegraphics[width = 0.42\columnwidth]{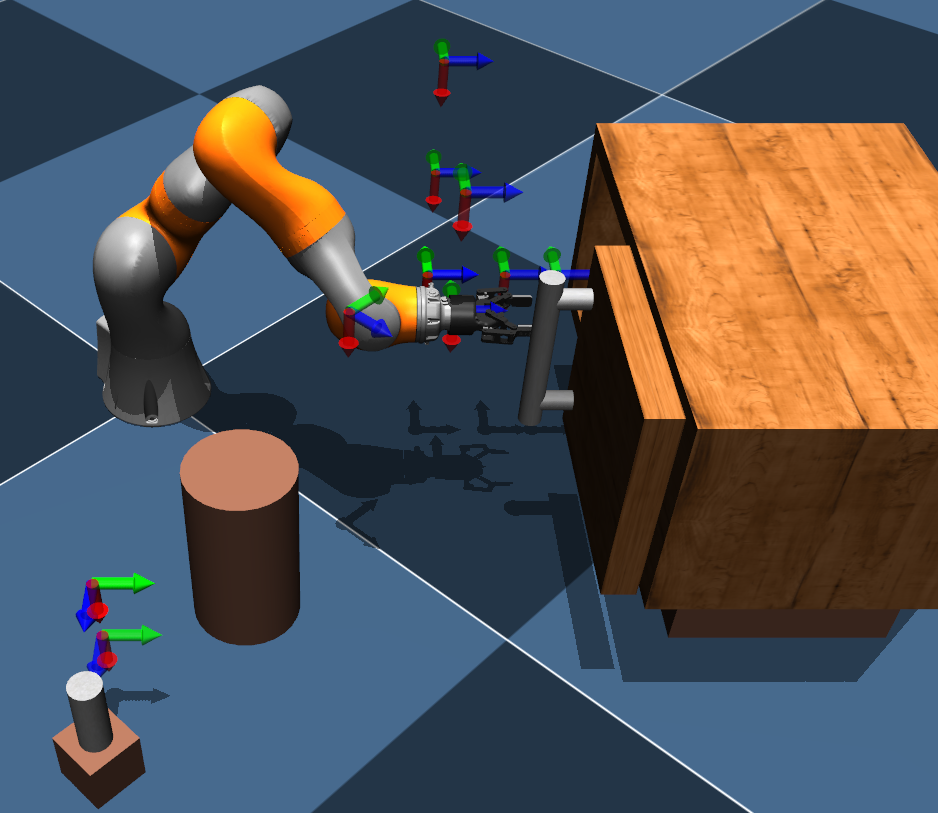}} \hfil
  \subfloat[]{\includegraphics[width = 0.42\columnwidth]{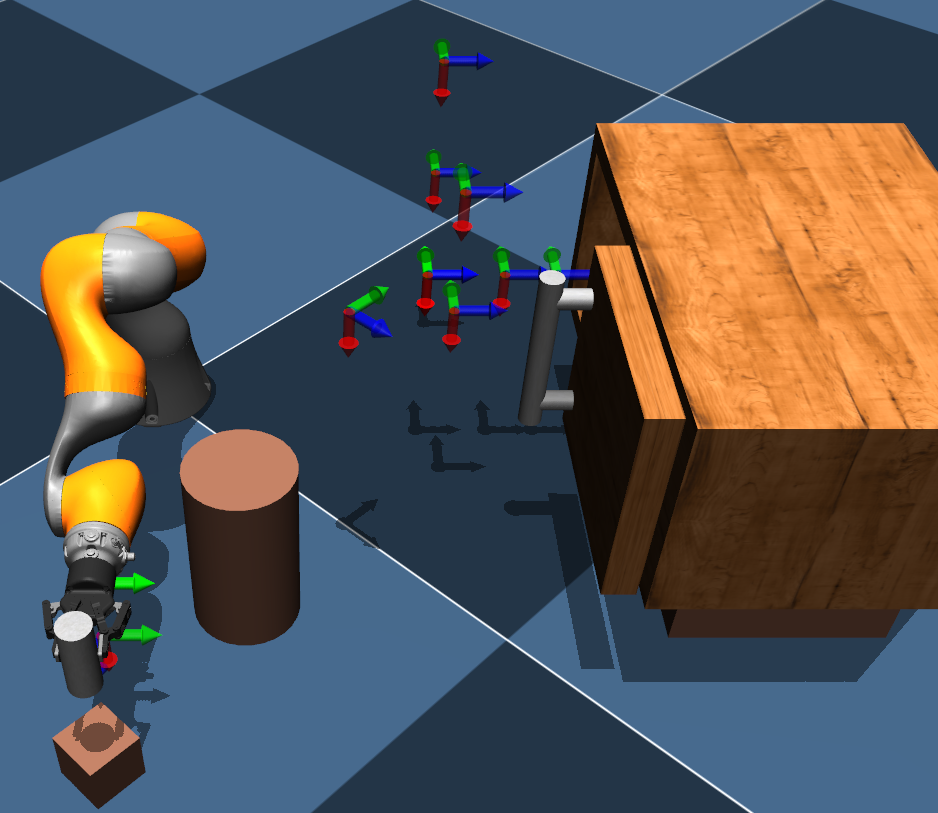}} \hfil
  \subfloat[]{\includegraphics[width = 0.42\columnwidth]{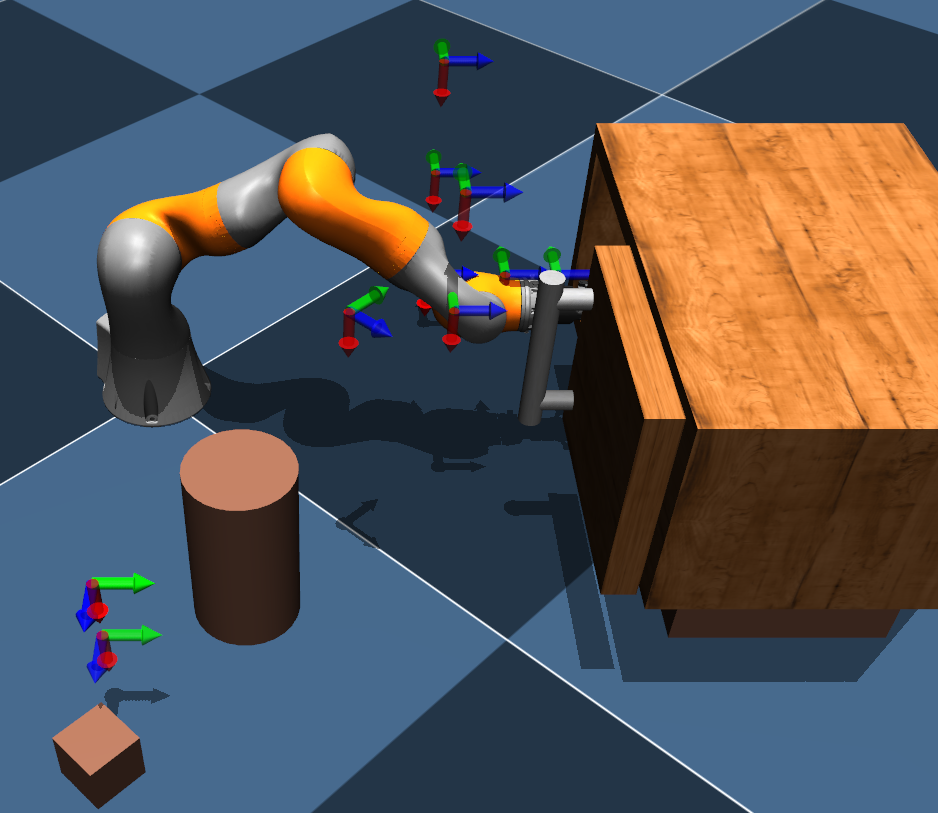}} \hfil
  \caption{\label{fig:sim} Sequential manipulation task in MuJoCo~\cite{Todorov2012}: The robot starts from an initial configuration in (a) and then moves through a sequence of waypoints to open the cabinet door in (b). Afterwards, the robot must avoid the cylindrical obstacle while approaching and grasping the object in (c). Finally, the robot places the object into the cabinet in (d).}
\end{figure}
\begin{table*}[ht]
    \centering
    \medskip
    \caption{Comparison with MoveIt~\cite{Coleman2014} Planners}
    \label{tab:comparison}
    \begin{tabular}{c|c |c|c|c |c|c|c |c|c|c}
        \hline
        & & \multicolumn{3}{c|}{Path Length [\si{\radian}]} & \multicolumn{3}{c|}{Trajectory Duration [\si{\second}]} & \multicolumn{3}{c}{Planning Time [\si{\second}]}\\
        Algorithm & Type & min & max & avg & min & max & avg & min & max & avg \\
        \hline
        \hline
        RRTConnect & offline & 31.0896 & 43.2777 & 36.3912 & 13.0539 & 21.1495 & 15.3991 & 0.0997 & 0.2355 & 0.1613\\
        T-RRT & offline & 29.7765 & 53.5128 & 32.8817 & 13.0094 & 20.1818 & 14.2923 & 0.1119 & 8.8188 & 0.8361\\
        RRT* & offline & 29.5847 & 34.8799 & 31.5359 & 13.0449 & 15.6965 & 14.0404 & 130.0704 & 130.1232 & 130.0838\\
        %WMPC (ours) & online & & & & & & & & & & \\
        WMPC (ours) & online & \multicolumn{3}{c|}{33.4563} & \multicolumn{3}{c|}{15.1} & \multicolumn{3}{c}{0.1}\\
        \hline
    \end{tabular}
\end{table*}
In the first scenario, the robot must move through several waypoints to solve a sequential manipulation task in simulation using MuJoCo~\cite{Todorov2012}, placing an object in a cabinet where the robot must open the door first.
This simulation experiment intends to assess the performance of the proposed online wMPC planner in terms of the resulting path length and trajectory duration compared to state-of-the-art sampling-based motion planners implemented in MoveIt~\cite{Coleman2014}.

In a second lab experiment, the proposed wMPC planner shows its unique feature to account dynamically for new and removed waypoints in real-time.
To this end, the robot must grasp a cylinder from a table and insert it into a cup. The cylinder and the cup can be moved, forcing the robot to replan dynamically. Waypoints determine the approach directions for the grasping and insertion motions.

%The parameters for the proposed planning algorithm used in the experiments, unless stated otherwise, are summarized in Table~\ref{tab:wmpc_params}, and the bounds for the optimization problem~(\ref{eqn:jerk_opt}) are given in Table~\ref{tab:bounds}.
Table~\ref{tab:wmpc_params} shows the parameters for the wMPC algorithm used in the experiments unless stated otherwise. Table~\ref{tab:bounds} gives the bounds for the optimization problem~(\ref{eqn:jerk_opt}).
Input bounds $\underline{\vect{u}}$, and $\overline{\vect{u}}$ are neglected because jerk is already regularized in the objective function~(\ref{eqn:opt_cost}), and the RRT-type algorithms used in Section~\ref{sec:sequential_manipulation} cannot account for them.
The optimization problem~(\ref{eqn:jerk_opt}) is implemented as a \ROS node~\cite{Quigley2009} in Python using CasADi~\cite{Andersson2019} and solved with the nonlinear interior point solver IPOPT~\cite{Waechter2006} and MA57. 
Planning times of $\SI{100}{\milli\second}$ are achieved, including the online solution for the analytic inverse kinematics~\cite{Shimizu2008} for new Cartesian waypoints and a desired goal.
Compatible inverse kinematics solutions for the waypoints are obtained by choosing the solution closest to the previous one in a least-squares sense.
%The planner receives the waypoints and the goal in Cartesian space as a position vector and a quaternion.
%The analytic inverse kinematics solution from~\cite{Shimizu2008} for the \kuka robot used in this work calculates the corresponding joint configurations.
For collision checking, the robot and the robot's gripper are approximated with spheres in the relevant locations.
%as depicted in Fig.~\ref{fig:gripper_collision}. 
% Table~\ref{tab:collision_objects} summarizes the sphere properties and their locations.
The collision object cylinders are modeled as capsules, and the ground plane is an additional obstacle restricting the motion in $z$-direction.
No collision checking is done for the cup in the dynamic replanning experiment because modeling the hollow object is more involved. 
Instead, waypoints are used to approach the cup from above, which ensures that no collision occurs with the cup.
A video of the presented scenarios and additional scenarios can be found at \url{www.acin.tuwien.ac.at/8a92}.

\subsection{First Scenario: Simulation Experiment for Sequential Manipulation}
\label{sec:sequential_manipulation}

In this simulation experiment, the ability of the proposed planning algorithm to pass several waypoints to achieve a sequential manipulation task is tested and compared to offline planning algorithms in MoveIt~\cite{Coleman2014} regarding path length and trajectory duration.
The robot must move through waypoints to first open a cabinet door.
Afterward, the robot must grasp a cylindrical object while avoiding an obstacle.
Finally, the object must be placed in the cabinet before the robot can retreat to its initial configuration again.
Fig.~\ref{fig:sim} shows the scene setup, including the waypoints.

In order to assess the performance of the proposed (local) online wMPC planner, the same scenario is solved using (global) offline sampling-based planners implemented in MoveIt~\cite{Coleman2014}, specifically RRTConnect~\cite{Kuffner2000}, RRT*~\cite{Karaman2011}, and T-RRT~\cite{Jaillet2010}.
A path segment is planned between each waypoint.
The same analytic inverse kinematics~\cite{Shimizu2008} solution is used to calculate the corresponding waypoints in the joint space, as in the presented wMPC approach.
A time parametrization is obtained for the entire path using the Time-Optimal Trajectory Generation~(TOTG) algorithm~\cite{Kunz2012}.
Table~\ref{tab:bounds} specifies the acceleration limits, and the velocity limits are halved to obtain meaningful interaction speeds.
%To achieve a comparable trajectory duration of the proposed approach, the scaling factor $\sigma$ is very large with $\sigma = 2000$.
The scaling factor is chosen as $\sigma = 2000$ to achieve a near-time-optimal behavior.
Furthermore, the collision avoidance cost is set to $w_3 = 10 \sigma$.

Table~\ref{tab:comparison} summarizes the results of the comparison. 
Due to the stochastic nature of the RRT-type planners, the results are averaged over 50 runs.
The reported planning time for the RRT-type planners includes the planning time for all path segments and the calculation of the time parametrization.
The results show that the proposed online wMPC planner achieves a path length close to the average of T-RRT, which does not quite reach as short paths as RRT* but is shorter on average than RRTConnect. 
%Regarding trajectory duration, the receding horizon approach achieves faster trajectories than the average duration of RRTConnect but slower trajectories compared to T-RRT and RRT*. 
The trajectory duration achieved by wMPC is slightly longer than the average duration achieved by RRT* and T-RRT and comparable to the average trajectory duration of RRTConnect.
However, the minimum duration is still shorter for RRTConnect, T-RRT, and RRTConnect, which is related to the smaller minimum path lengths for these approaches.
One reason for the longer trajectory duration of the proposed wMPC approach is that by minimizing the jerk, trajectories are smoother.
While the proposed approach is permanently restricted to \SI{0.1}{\second} planning time, RRTConnect is the only algorithm that does not exceed this planning time in rare cases.
RRT* is looking for an asymptotically optimal solution and is planning until the time limit of \SI{10}{\second} per path segment is reached.
%In this scenario, T-RRT has a success rate of only \SI{80}{\percent}.
%RRTConnect and RRT* always found a solution. 
%In contrast, the proposed online wMPC approach is deterministic.

The results show that the proposed wMPC approach can successfully plan in real-time through the desired waypoints with a receding horizon while still obtaining good performance in path length and trajectory duration compared to the full-horizon RRT-type planners in this scenario.
Compared to the sampling-based planners, the proposed approach is susceptible to local minima due to the nonlinear optimization and the receding horizon.
Hence, wMPC might fail to find a suitable solution for more cluttered scenes.
However, due to the possibility of incorporating the waypoints, the planning problem can often be significantly simplified by intelligent task planning and waypoint placement.
The main advantage of the proposed approach is that kinematic and dynamic constraints, in addition to waypoints, can systematically be considered in the optimization problem while planning over a receding horizon to keep planning times low.

%\begin{figure}[!t]
%    \centering
%    \includegraphics[width=0.8\columnwidth]{graphics/sim/start.png}
%    \caption{Sequential manipulation task in MuJoCo~\cite{Todorov2012}.}
%    \label{fig:sim}
%\end{figure}

\subsection{Second Scenario: Lab Experiment for Dynamic Replanning and Reactive Behavior}
\label{sec:dynamic_replanning}

\begin{figure}[!t]
  \centering
  \subfloat[]{\includegraphics[width = 0.42\columnwidth]{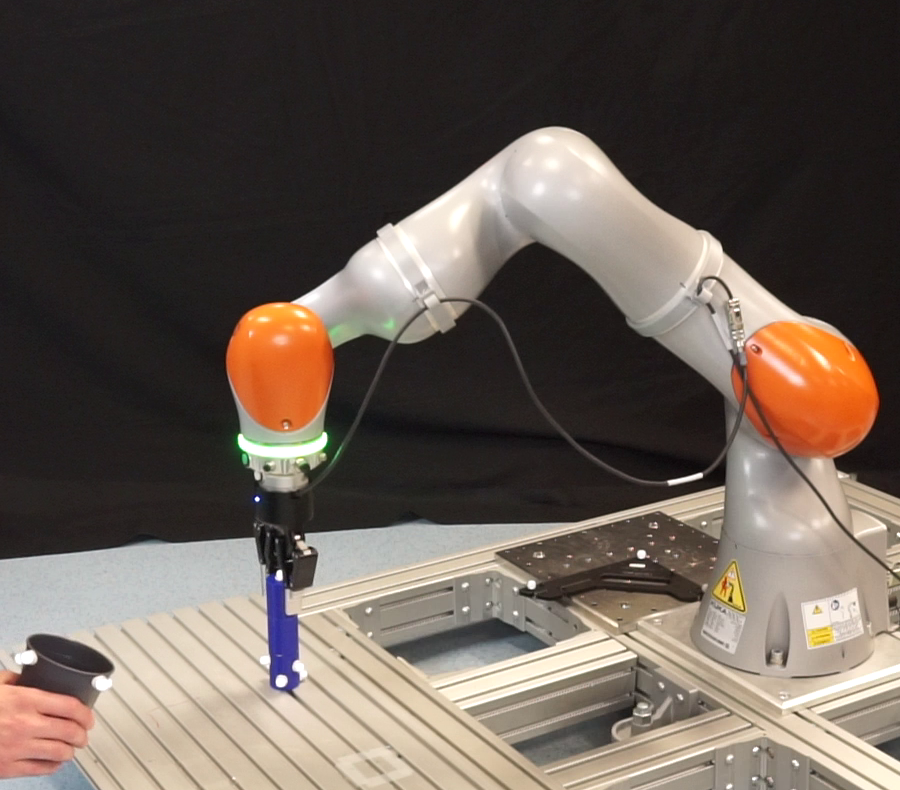}} \hfil
  \subfloat[]{\includegraphics[width = 0.42\columnwidth]{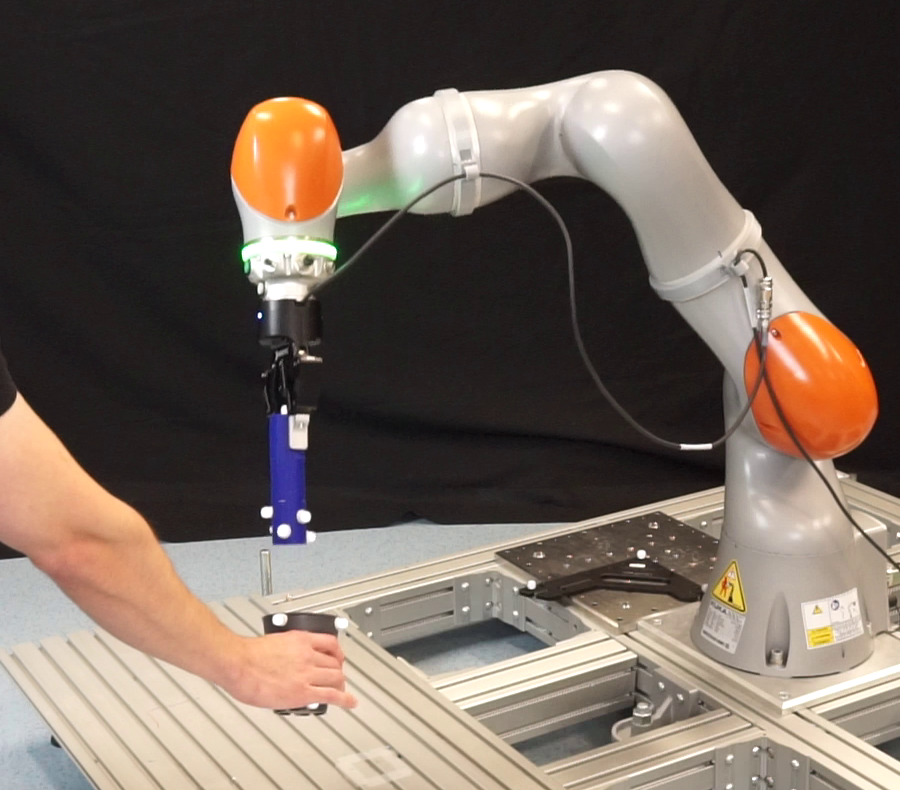}} \hfil
  \subfloat[]{\includegraphics[width = 0.42\columnwidth]{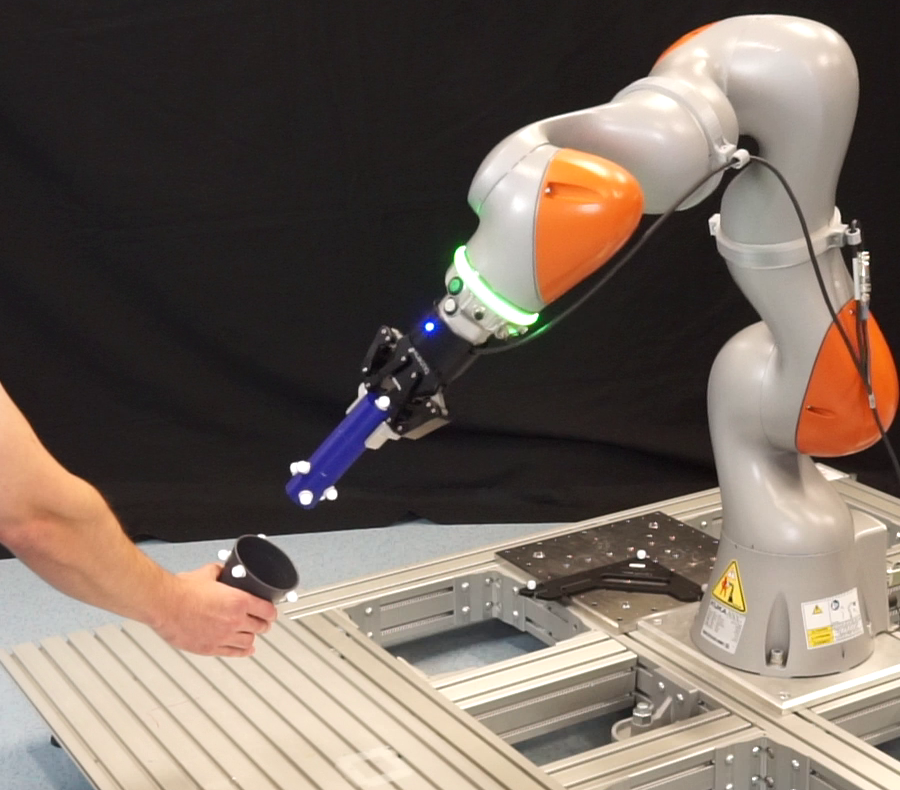}} \hfil
  \subfloat[]{\includegraphics[width = 0.42\columnwidth]{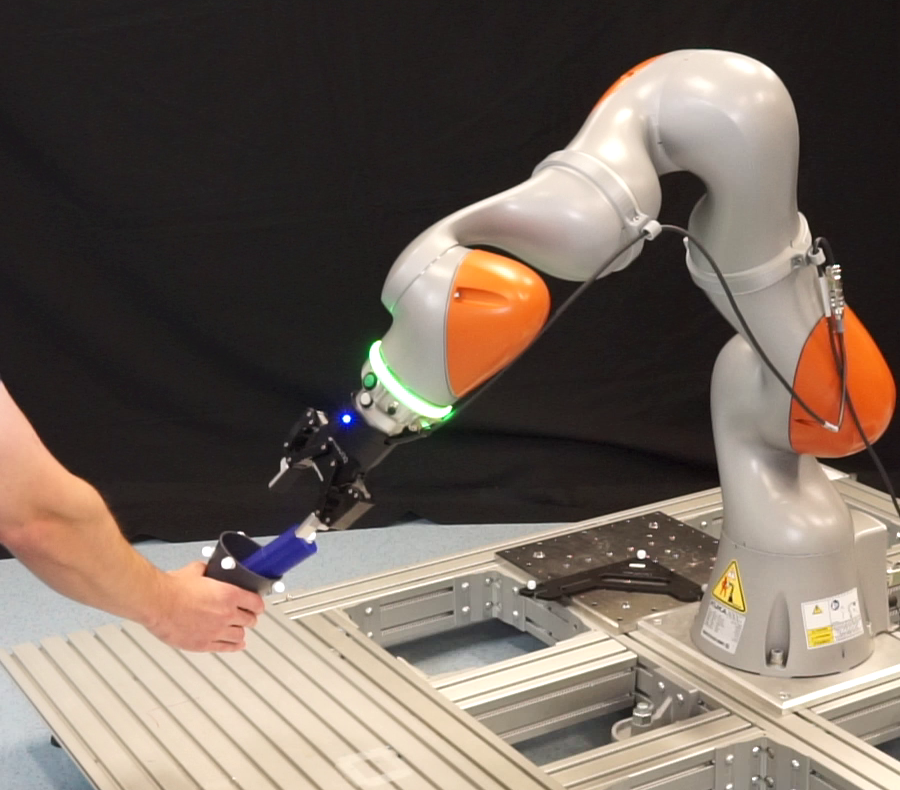}} \hfil
  \caption{\label{fig:title_overlay} Robotic grasping scenario with waypoints and dynamic replanning: The robot grasps the cylinder \circled{3} in (a) after passing through a waypoint \circled{2} above it. The cup is approached in (b) through a waypoint \circled{4} to align the approach direction. After moving the cup, the robot adjusts the waypoint \circled{5} and the goal \circled{6} for the new cup position (c) and places the object in (d).}
\end{figure}

In this lab experiment, the robot must grasp a cylinder with a height of $h = \SI{0.15}{\metre}$ and a radius of $r = \SI{0.02}{\metre}$ and place it into a cup.
The locations of the cylinder and the cup are tracked using \Optitrack with markers placed on their surface.
Fig.~\ref{fig:title_overlay} shows an overview of the experimental setup and the scenario sequence \circled{1} - \circled{6} executed by the robot.
%\begin{figure*}[!t]
%  \centering
%  \subfloat[]{\includegraphics[width = 0.49\columnwidth]{graphics/frames/cylinder_grasp.png}} \hfil
%  \subfloat[]{\includegraphics[width = 0.49\columnwidth]{graphics/frames/approach_1.png}} \hfil
%  \subfloat[]{\includegraphics[width = 0.49\columnwidth]{graphics/frames/approach_2.png}} \hfil
%  \subfloat[]{\includegraphics[width = 0.49\columnwidth]{graphics/frames/cylinder_in_cup.png}} \hfil
%  \caption{\label{fig:title_overlay} Robotic grasping scenario with waypoints and dynamic replanning. The robot grasps the cylinder \circled{3} in (a) after passing through a waypoint \circled{2} above it. The cup is approached in (b) through a waypoint \circled{4} to align the approach direction. After moving the cup, the robot adjusts the waypoint \circled{5} and the goal \circled{6} for the new cup position (c) and places the object in (d).}
%\end{figure*}
A simple task planner ensures good approach directions for the grasp and placement by placing waypoints \SI{0.1}{\metre} and \SI{0.15}{\metre} above the objects, respectively.

%\begin{figure}
%  \centering
%  \begin{tikzpicture}
%    \node[inner sep = 0pt] (robot_image) at (0, 0) {\includegraphics[width=\columnwidth]{graphics/setup.jpg}};
%  \end{tikzpicture}
%  \caption{\label{fig:setup} This figure depicts the experimental setup. The \kuka robot is equipped with a \robotiq gripper. The cylinder and the cup are tracked using \Optitrack cameras, including the three cameras in the figure and two additional ones on the opposite side.}
%\end{figure}

\begin{figure}[t]
  \centering
  \medskip
  \includegraphics{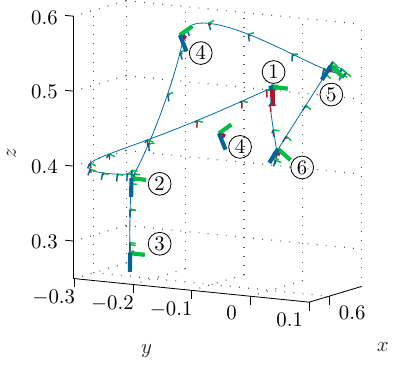}
  \caption{\label{fig:replanning_3d_traj}Cartesian end-effector trajectory for the dynamic replanning experiment: The robot starts at \circled{1}, moves towards the cylinder through a waypoint at \circled{2}, and grasps the cylinder at \circled{3}. Afterward, the robot moves back through \circled{2} and attempts to put the cylinder in the cup at \circled{4}, moving to the appropriate waypoint. However, the cup is moved, and the robot adjusts the trajectory to move through a waypoint at \circled{5} and places the cylinder in the cup at \circled{6}. Finally, the robot returns to the initial pose at \circled{1}.}
\end{figure}

\begin{figure*}[t]
  \centering
  \medskip
  \includegraphics{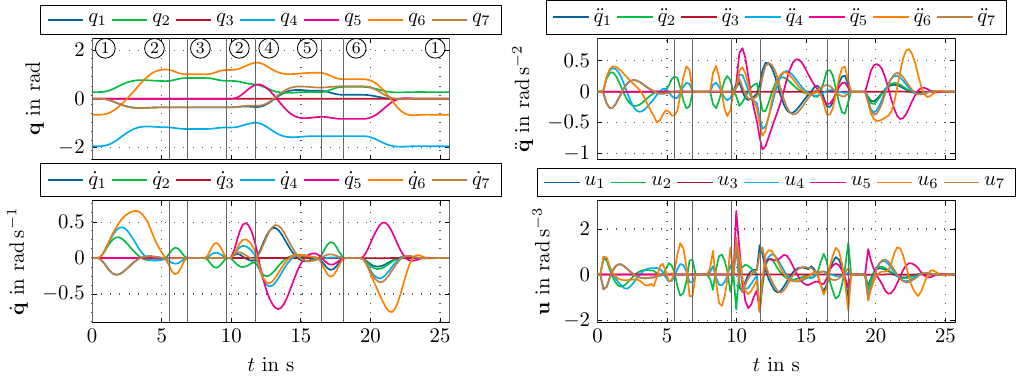}
  \caption{\label{fig:replanning_plot} Planned joint-space trajectories for the dynamic replanning scenario.}
\end{figure*}

A joint-space inverse dynamics control law follows the planned trajectory after interpolating it using first-order-hold according to (\ref{eqn:sys_d}) to adapt to the higher rate of the control law.
Fig.~\ref{fig:replanning_3d_traj} shows the Cartesian end-effector trajectory, and Fig.~\ref{fig:replanning_plot} depicts the corresponding motion in the joint space.
The robot moves through the waypoint \circled{2} to grasp the cylinder \circled{3}.
One can observe that the motion is smooth throughout the waypoint to reach the goal.
Similarly, when the robot approaches the final pair of waypoint \circled{5} and goal \circled{6}, the robot passes smoothly through the waypoint without stopping.
The smoothness and continuous motion are due to the split-horizon formulation of the optimization problem~(\ref{eqn:jerk_opt}), which simultaneously optimizes the movement through the waypoint and the motion to the goal.
The cup is moved by hand between the retreating waypoint \circled{2} and the waypoint for the placement \circled{5}. 
Therefore, the algorithm has to replan several times to adjust to a new waypoint and a new goal generated by the vision system.
Nevertheless, the resulting motion remains smooth between \circled{2} and \circled{5} in Fig.~\ref{fig:replanning_plot}.
One of the waypoints and the corresponding goal while moving the cup are shown at \circled{4}, where the robot attempts to place the cylinder in the cup before the cup is moved again, requiring the algorithm to replan for the final waypoint \circled{5} and goal \circled{6}.

%% file: sections/conclusion.tex
This work presents a novel waypoint model predictive control (wMPC) approach for systematically incorporating dynamically changing waypoints into a receding horizon trajectory optimization. When a waypoint becomes reachable within the optimization horizon, it is added to the optimization problem as a constraint. This way, the waypoint is passed with a certain tolerance but without necessarily stopping there. This approach enables dynamic replanning in real-time and reactive tracking of waypoints, which may result from superordinate task planning algorithms. 
Simulation results show that the proposed (local) real-time receding horizon approach yields path lengths and trajectory durations in a sequential manipulation task similar to (global) sampling-based RRT-type planners, however, with online capability. Furthermore, experimental results on a \kuka robot demonstrate the reactive online replanning capabilities of the proposed algorithm, see the video in \url{www.acin.tuwien.ac.at/8a92}. 

In future work, finding waypoints for sequential manipulation tasks in a dynamically changing scene and utilizing the replanning capabilities of the wMPC algorithm to adapt to changes and feedback from the environment will be further explored.